\documentclass[conference]{IEEEtran}

\usepackage{cite}
\usepackage[hidelinks,hypertexnames=false]{hyperref}
\usepackage{amsmath,amssymb,amsfonts}

\usepackage{graphicx}
\usepackage{algorithmic}
\usepackage{textcomp}
\usepackage[table,xcdraw]{xcolor}
\usepackage{tikz}
\usetikzlibrary{positioning,arrows.meta}
\usepackage{booktabs}
\usepackage{stfloats}
\usepackage{tabularx}
\usepackage{adjustbox}
\usepackage{float}
\usepackage{placeins}
\usepackage{tcolorbox}
\usepackage{colortbl}
\usepackage{multirow}
\usepackage{url}
\usepackage{balance}
\hypersetup{
    colorlinks=false
}
\def\BibTeX{{\rm B\kern-.05em{\sc i\kern-.025em b}\kern-.08em
    T\kern-.1667em\lower.7ex\hbox{E}\kern-.125emX}}
\begin{document}

\definecolor{figyellow}{RGB}{255,242,204}
\definecolor{figblue}{RGB}{221,235,247}
\definecolor{figgreen}{RGB}{226,239,218}
\definecolor{figpurple}{RGB}{234,209,220}
\definecolor{figred}{RGB}{244,204,204}

\title{
Comparative Performance and Parameter-Efficient Adaptation of DINOv2 for Active Trachoma Classification}

\author{
 \IEEEauthorblockN{Kibrom Gebremedhin}
\IEEEauthorblockA{\textit{Department of Computer Science} \\
\textit{Mekelle University}\\
Mekelle, Ethiopia \\
kibrom.gebremedhin@mu.edu.et}
\and
\IEEEauthorblockN{Hadush Hailu}
\IEEEauthorblockA{\textit{Department of Computer Science} \\
\textit{Maharishi International University}\\
Fairfield, IA, USA \\
hadush.gebrerufael@miu.edu}
\and
\IEEEauthorblockN{Bruk Gebregziabher}
\IEEEauthorblockA{\textit{Signal Technologies}\\
Germany \\
bruk@signaltech.xyz}
\and
\IEEEauthorblockN{Yordanos Hailu}
\IEEEauthorblockA{\textit{Department of Computer Science} \\
\textit{MicroLink Information Technology College}\\
hailuyordanos061@gmail.com}
}

\maketitle

\begin{abstract}
Automated grading of conjunctival photographs could reduce the cost and variability
of trachoma prevalence surveys, but the relative value of modern pretrained visual
representations, lightweight feature adaptation, and training-objective design has
not been established under a common protocol. This study presents a controlled
evaluation for binary classification of Trachomatous Inflammation--Follicular (TF)
versus Normal using 1,546 images from the public UCSF/Lietman collection. Images are
processed using the OPTED pipeline for zero-shot tarsal-conjunctiva segmentation,
alignment, cropping, and standardization. We first compare six pretrained backbones
using a common classification pipeline and then evaluate four lightweight adaptation
mechanisms on DINOv2 ViT-B/14. Under stratified five-fold cross-validation, DINOv2
with Efficient Channel Attention (ECA) and focal-plus-center loss achieved
$91.66\pm0.97\%$ accuracy, $90.69\pm1.10\%$ macro-F1, and
$96.06\pm0.71\%$ AUC. ECA introduces only five learnable parameters while matching
the performance of substantially larger alternatives. Objective ablation further
showed that ECA did not consistently improve plain DINOv2 across loss functions; the
lowest-variance 91.66\% accuracy was obtained with cross-entropy plus center loss.
Overall, the fine-tuned DINOv2 representation provided most of the predictive
performance, while ECA offered a highly parameter-efficient refinement whose effect
depended on the training objective. The resulting workflow provides a reproducible
benchmark for active trachoma image classification.
\end{abstract}


\section{Introduction}

Trachoma, caused by ocular infection with \emph{Chlamydia trachomatis}, is the
world's leading infectious cause of blindness, affecting approximately
1.9~million people and remaining a public health problem in
30~countries \cite{who2025trachoma}. Active disease disproportionately affects children in low-resource settings, while its blinding sequelae-trichiasis and corneal opacity-manifest predominantly in adults following decades of repeated infection.

The global burden is heavily concentrated in Sub-Saharan Africa. Ethiopia alone
harbours more than 50\% of the worldwide burden of active trachoma, with regional
TF prevalence estimates reaching 26.1\% in Tigray and endemic conditions persisting
across Amhara, Oromia, and SNNP regions \cite{solomon2022}. Despite the WHO's
SAFE strategy (Surgery, Antibiotics, Facial cleanliness, Environmental improvement)
and a global elimination target of 2030 \cite{who2021}, progress remains hampered
by limited diagnostic infrastructure, a shortage of trained graders, and significant
inter-observer variability in clinical assessment. Studies report inter-grader
Cohen's $\kappa$ values as low as 0.44 for TF \cite{kim2019}, introducing systematic
underreporting in prevalence surveys.

The WHO simplified trachoma grading system classifies active disease into
Trachomatous Inflammation Follicular (TF) and Trachomatous Inflammation Intense
(TI), distinguished by the presence and distribution of follicles on the tarsal
conjunctiva of the everted upper eyelid \cite{thylefors1987}. Automated,
objective, and scalable trachoma classification using deep learning offers a viable
pathway to overcoming these diagnostic barriers. CNN-based systems have demonstrated
performance approaching that of certified graders on controlled datasets
\cite{kim2019,socia2022,milad2023, pan2024,joye2025, yenegeta2023, zewudie2024adaptive}, and recent advances in self-supervised
vision transformers have opened new possibilities for robust medical image analysis
with limited labelled data.

Despite this progress, two critical gaps remain. First, no systematic comparison
of modern self-supervised vision transformers exists on the established Kim et al.\
benchmark \cite{kim2019}. Second, the interaction between backbone representation
quality and lightweight feature selection mechanisms has not been characterised for
conjunctival image analysis. This paper addresses both gaps through a controlled,
fully reproducible experimental study. Our key contributions are:

\begin{itemize}
  \item The first systematic comparison of six modern pretrained backbones---DINOv2
  ViT-B/14, EVA02-Base, ConvNeXt-Tiny, ConvNeXtV2-Base, BiomedCLIP, and
  MobileNetV3-Large---on the Kim et al.\ trachoma benchmark under identical
  experimental conditions.
  \item A feature selector ablation evaluating SE-Gate (74,544\,params), L0 Hard
  Concrete (768\,params), ECA-Net (5\,params), and ECA+STG (773\,params) on the
  frozen DINOv2 backbone.
  \item Demonstration that a five-parameter ECA-Net gate achieves higher AUC-ROC
  than a 74,544-parameter SE-Gate, advancing the case for minimal-complexity
  attention mechanisms in clinical imaging pipelines.
  \item A fully reproducible end-to-end workflow integrating OPTED SAM\,3
  zero-shot segmentation with downstream classification, establishing new
  benchmark results.
\end{itemize}

\section{Related Work}

Kim et al.\ established the public conjunctival-image benchmark and showed that a
CNN could approach expert TF grading \cite{kim2019}. Subsequent work explored
ResNet-101 \cite{socia2022}, automated model selection \cite{milad2023},
SAM-assisted ROI extraction \cite{pan2024}, field-scale CNN deployment
\cite{joye2025}, and proprietary-data models \cite{yenegeta2023}. Attention-based
VGG16 \cite{zewudie2024adaptive} and hybrid CNN--transformer models
\cite{zewudie2024pvit} further improved feature modeling, but these studies used
different datasets, splits, labels, or objectives and did not compare modern
self-supervised backbones under one protocol.

ViT introduced patch-token self-attention \cite{dosovitskiy2021}; DINOv2 later
combined self-distillation and masked-image learning at large scale
\cite{oquab2024}. EVA02 provides another masked-image transformer
\cite{fang2023}, BiomedCLIP uses biomedical image--text pretraining
\cite{zhang2023}, and ConvNeXt/ConvNeXtV2 retain convolutional inductive biases
within modernized designs \cite{liu2022}. These alternatives motivate the
controlled backbone comparison in this study.

For adaptation, SENet uses bottleneck channel recalibration \cite{hu2018}, while
ECA replaces that bottleneck with a low-cost one-dimensional convolution
\cite{wang2020}. L0 Hard Concrete \cite{louizos2018} and stochastic gates
\cite{yamada2020} instead learn sparse selectors. Their relative value for
conjunctival representations has not previously been tested systematically.

\begin{figure*}[t]
  \centering
  \includegraphics[width=\textwidth]{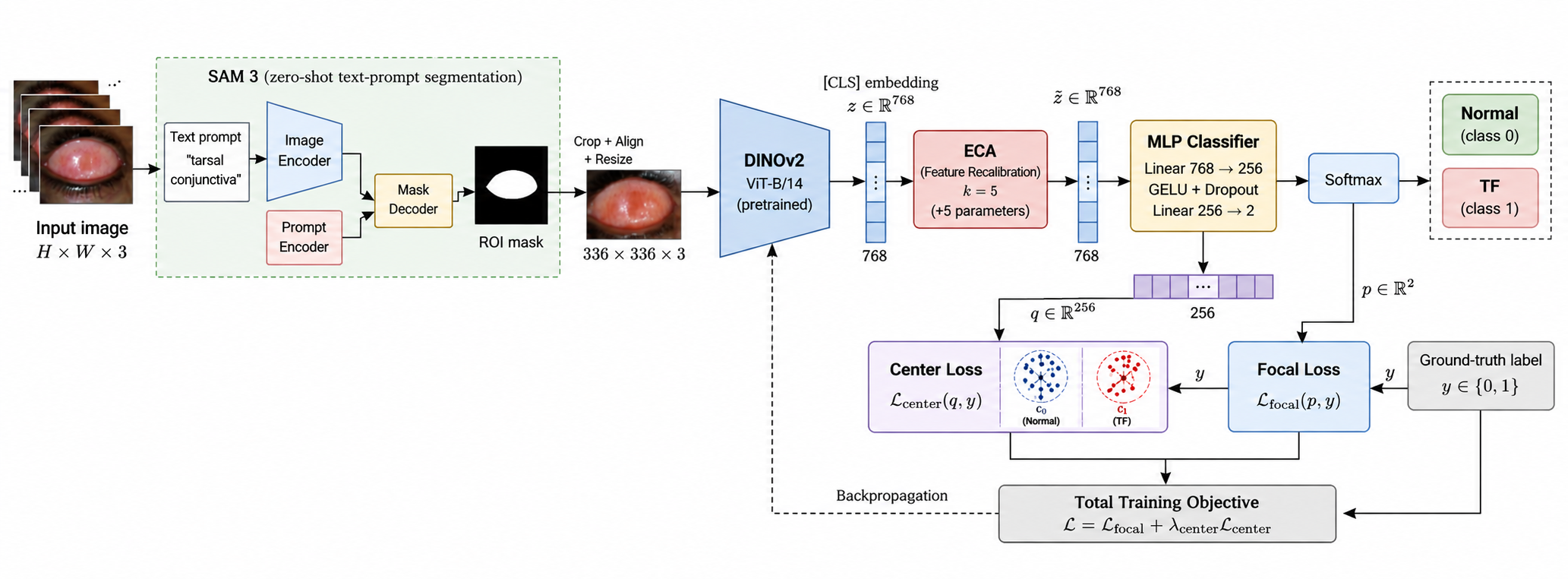}
  \caption{Proposed framework: SAM~3 ROI localization, DINOv2 feature
  extraction, ECA recalibration, and MLP classification with focal and center
  losses.}
  \label{fig:pipeline}
\end{figure*}

\section{Methodology}
\subsection{Study Design and Research Questions}

We used a controlled comparative design to separate three potential sources of
performance: the pretrained visual backbone, the feature adaptation module, and
the training objective. The study addressed three questions: (RQ1) whether
DINOv2 provides a stronger representation for active trachoma classification
than alternative pretrained CNN, transformer, and biomedical vision--language
backbones; (RQ2) how much task-specific selection improves on an identity
(no-selector) DINOv2 baseline, and at what parameter cost; and (RQ3) whether the
effect of ECA depends on focal and center-loss supervision. The same binary task,
fold membership, model-selection rule, and test metrics were retained across the
corresponding comparisons. Published trachoma systems are considered separately
as contextual benchmarks because their data partitions and evaluation protocols
are not necessarily identical to ours.

\subsection{Dataset and Task Definition}

Experiments used the public UCSF/Lietman dataset released by Kim et al.\
\cite{kim2019}. It contains 1,656 photographs of everted upper eyelids from the
PRET and TANA~II studies in Ethiopia and Niger: 1,019 Normal, 365 TF, 110 TI, and
162 combined TF--TI images. We defined active follicular inflammation as the
positive class by combining TF and TF--TI, and excluded TI-only images. The final
binary dataset therefore comprised 1,546 images: 1,019 Normal (65.9\%) and 527 TF
(34.1\%). The resulting 1.93:1 class ratio motivated the evaluation of a
class-weighted focal objective in addition to ordinary cross-entropy. Patient
identifiers were not supplied with the released image-level metadata; consequently,
the folds were stratified by image and patient-level separation could not be
verified.

\subsection{OPTED Region-of-Interest Preprocessing}

Raw photographs include variable amounts of skin, eyelashes, gloved fingers,
illumination artefacts, and other content outside the tarsal conjunctiva. We used
the OPTED pipeline \cite{gebremedhin2025} to apply SAM~3 zero-shot text-prompt
segmentation of the inner eyelid surface, retain the highest-scoring mask, remove
the non-ROI background, crop the mask bounding box with 5\% padding, and align the
crop horizontally. Pixels outside the mask were set to zero; no mean-filled
background was used. Crops were stored both at their aligned source resolution and
as $224\times224$ PNG images obtained with LANCZOS interpolation. The classification
loader subsequently resized each image to the input resolution configured for its
backbone; DINOv2 ViT-B/14 used $336\times336$ inputs. SAM~3 was used only for
anatomical localization and did not receive disease labels.

Training images underwent random resized cropping (scale 0.8--1.0; aspect ratio
0.9--1.1), horizontal and vertical flips ($p=0.5$ each), rotation up to $15^\circ$
($p=0.5$), color jitter ($p=0.5$), Gaussian or median blur ($p=0.2$), and coarse
dropout of up to eight $16\times16$ regions ($p=0.3$). After the frozen-backbone
phase, Mixup ($\alpha=0.3$) or CutMix ($\alpha=1.0$) was selected on 50\% of
minibatches. Validation and test images received only resizing and ImageNet
normalization (mean $[0.485,0.456,0.406]$, standard deviation
$[0.229,0.224,0.225]$).

\subsection{Unified Model Formulation}

For an input image $x$, a backbone $f_{\theta}$ produces a representation $z$,
an optional adaptation module $g_{\phi}$ recalibrates it, and a classifier
$h_{\psi}$ produces two logits:
\begin{equation}
 z=f_{\theta}(x), \qquad \tilde z=g_{\phi}(z), \qquad
 \hat y=h_{\psi}(\tilde z).
 \label{eq:framework}
\end{equation}
For the identity baseline, $g_{\phi}(z)=z$. Gated variants use
$\tilde z=z\odot a_{\phi}(z)$, where $a_{\phi}(z)\in[0,1]^d$ and $\odot$
denotes element-wise multiplication. This common formulation permits a direct
comparison between the information already present in DINOv2 and the incremental
effect of task-specific adaptation.

Figure~\ref{fig:pipeline} summarizes the DINOv2 configuration. DINOv2 ViT-B/14
\cite{oquab2024} maps a $336\times336$ image to a 768-dimensional \texttt{[CLS]}
representation. The shared classification head applies
$\mathrm{LN}\rightarrow\mathrm{Dropout}(0.1)\rightarrow\mathrm{Linear}(768,256)
\rightarrow\mathrm{GELU}\rightarrow\mathrm{Dropout}(0.1)
\rightarrow\mathrm{Linear}(256,2)$. It contains 198,914 parameters, and its
256-dimensional hidden activation is used when center loss is enabled.

\subsection{Controlled Experimental Comparisons}

\subsubsection{Backbone Comparison}

We compared DINOv2 ViT-B/14, EVA02-Base \cite{fang2023}, ConvNeXt-Tiny and
ConvNeXtV2-Base \cite{liu2022}, BiomedCLIP \cite{zhang2023}, and
MobileNetV3-Large. These models span self-supervised and masked-image-modelled
transformers, modern convolutional networks, biomedical image--text pretraining,
and a mobile architecture. Each backbone used its configured input size
(DINOv2, ConvNeXt-Tiny, ConvNeXtV2, and MobileNetV3: 336; EVA02: 448;
BiomedCLIP: 224) and was coupled to ECA, the same classification head, training
schedule, folds, and evaluation procedure. This experiment isolates backbone
choice while respecting the input resolution of each pretrained implementation.

\subsubsection{Identity and Feature Adaptation}

With DINOv2 fixed as the backbone architecture, we compared an identity mapping
with four adaptation mechanisms. SE-Gate \cite{hu2018} uses a bottleneck MLP;
L0 Hard Concrete \cite{louizos2018} learns stochastic $\ell_0$-regularized
gates; ECA \cite{wang2020} applies a bias-free one-dimensional convolution of
kernel size five followed by a sigmoid; and ECA--STG combines ECA with stochastic
gates \cite{yamada2020}. Table~\ref{tab:selectors} shows the additional learnable
parameters. ECA is treated as low-capacity feature recalibration; no semantic or
spatial ordering is assumed for adjacent dimensions of the DINOv2 embedding.

\begin{table}[!t]
\caption{Adaptation mechanisms applied to DINOv2.}
\label{tab:selectors}
\centering
\footnotesize
\begin{tabular}{lrl}
\toprule
\textbf{Adapter} & \textbf{Parameters} & \textbf{Mechanism} \\
\midrule
Identity & 0 & No recalibration \\
ECA & 5 & Input-dependent soft gate \\
L0 Hard Concrete & 768 & Sparse stochastic gate \\
ECA--STG & 773 & ECA plus stochastic gate \\
SE-Gate & 74,544 & Bottleneck MLP gate \\
\bottomrule
\end{tabular}
\end{table}

\subsubsection{Objective Ablation}

The objective study crossed identity or ECA adaptation with cross-entropy or
class-weighted focal loss and with the presence or absence of center loss. For
target probability $p_{y_i}$, focal loss was
\begin{equation}
\begin{aligned}
 \mathcal{L}_{\mathrm{focal}}&=-\frac{1}{N}\sum_{i=1}^{N}
 \alpha_{y_i}(1-p_{y_i})^{\gamma}\log p_{y_i},\\
 \alpha&=[0.35,0.65],\qquad \gamma=2,
\end{aligned}
 \label{eq:focal}
\end{equation}
with label smoothing 0.05. Center loss operated on the 256-dimensional hidden
embedding $q_i$:
\begin{equation}
 \mathcal{L}_{\mathrm{center}}=\frac{1}{2N}\sum_{i=1}^{N}
 \lVert q_i-c_{y_i}\rVert_2^2,
 \label{eq:center}
\end{equation}
where $c_{y_i}$ is the centroid of class $y_i$. It reduces within-class
dispersion and was weighted by $\lambda_{\mathrm{center}}=0.005$; the
classification objective remains responsible for learning the decision boundary.
The general training objective was
\begin{equation}
 \mathcal{L}=\mathcal{L}_{\mathrm{cls}}+
 \lambda_{\mathrm{center}}\mathcal{L}_{\mathrm{center}}+
 \mathcal{L}_{\mathrm{gate}},
 \label{eq:objective}
\end{equation}
where disabled terms are zero and $\mathcal{L}_{\mathrm{gate}}$ is present only
for the L0 and STG-based selectors. The ablation comprised DINOv2 without feature adaptation with
focal-plus-center or cross-entropy, and DINOv2--ECA with focal-plus-center,
cross-entropy-plus-center, focal without center loss, or cross-entropy alone.

\subsection{Training and Evaluation Protocol}

We used five-fold stratified cross-validation generated with seed 123. Each fold
held out approximately 20\% for testing; the remaining 80\% was divided
stratifiably into approximately 90\% training and 10\% validation data, producing
1,113--1,114 training, 123 validation, and 309--310 test images per fold. Saved
fold assignments were reused for the backbone, selector, and objective studies.
Training randomness was initialized with seed $123+k$ for fold $k$.

Training proceeded for at most 60 epochs. During epochs 1--8, the pretrained
backbone was frozen and the adapter and head were optimized at $10^{-3}$. At
epoch 9 the complete backbone was unfrozen; AdamW then used $5\times10^{-6}$ for
the backbone and $5\times10^{-5}$ for the adapter, head, and center parameters,
with weight decay 0.01 and per-step cosine decay to $10^{-7}$. The batch size was
8, gradients were clipped to norm 1.0, mixed-precision training was used, and an
exponential moving average with decay 0.999 supplied validation and test weights.
Early stopping with patience 20 monitored validation macro F1 and restored the
checkpoint with the highest validation macro F1. Because the backbone was unfrozen after epoch 8, our
representation-sufficiency analysis concerns a fine-tuned, rather than frozen,
DINOv2 representation.

For each held-out fold we measured accuracy, macro F1, ROC AUC, TF sensitivity,
and Normal specificity. Results are reported as the mean $\pm$ standard deviation
over the five test folds. AUC provides a threshold-independent measure, whereas
sensitivity and specificity expose the two clinically different error types
\cite{hanley1982,kocak2025}. Differences are described as observed mean effects;
no claim of statistical significance is made without a paired inferential test.

\begin{figure}[H]
\centering
\includegraphics[width=\columnwidth]{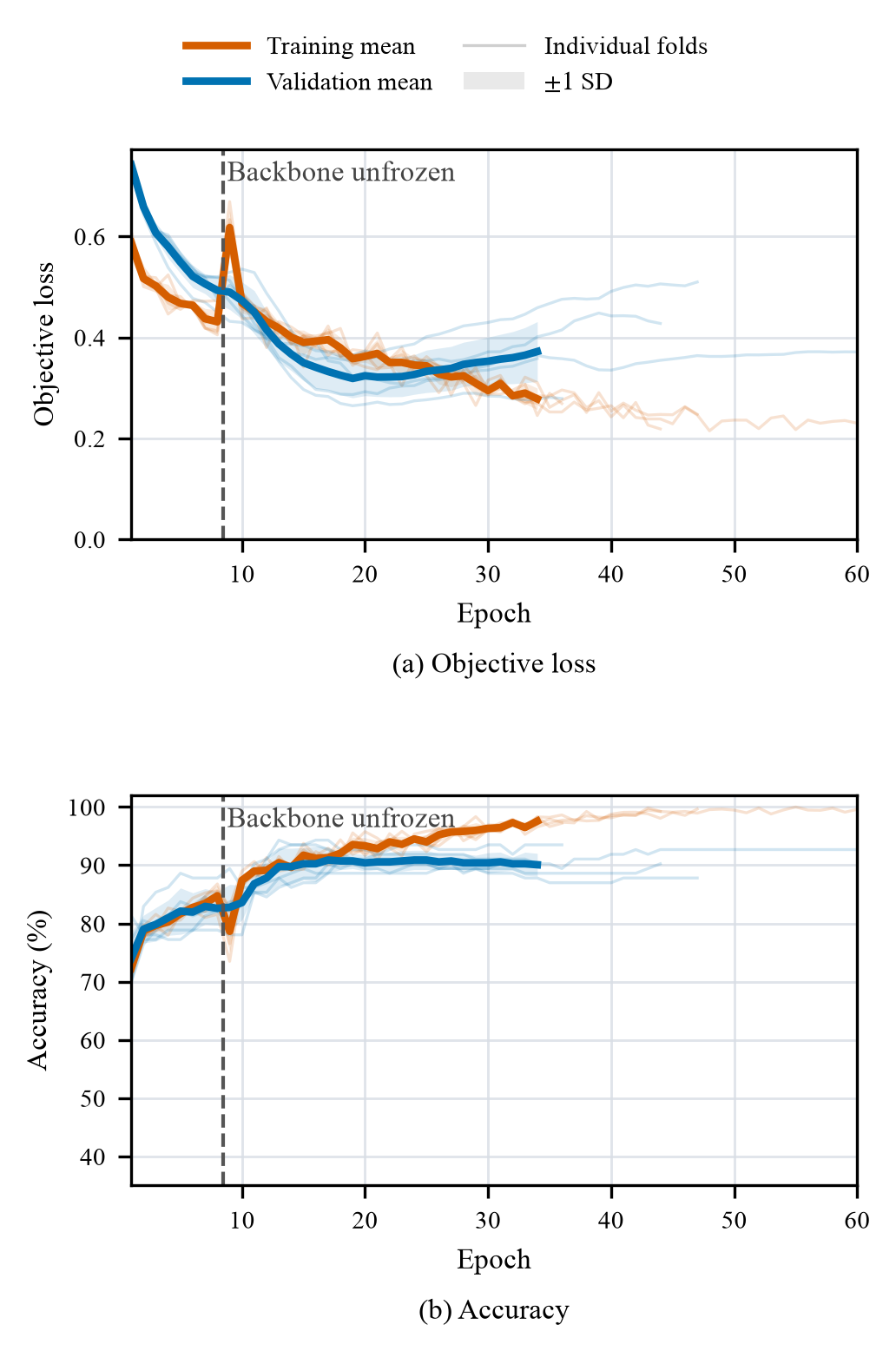}
\caption{Five-fold training dynamics: (a) objective loss and (b) accuracy.
Bold curves are means, bands show $\pm1$ SD, and the dashed line marks backbone
unfreezing.}
\label{fig:training_dynamics}
\end{figure}

Figure~\ref{fig:training_dynamics} shows the optimization dynamics for the
DINOv2--ECA cross-entropy-plus-center configuration, which attained the highest
mean accuracy with the lowest accuracy variation in
Table~\ref{tab:objective_ablation}. The temporary change at epoch~9 coincides
with backbone unfreezing and the optimizer reset to the lower fine-tuning
learning rates. Thereafter, training loss continued to decrease and training
accuracy approached 100\%, while the validation curves stabilized near 90\%
accuracy and began to separate from the training curves. This pattern supports
checkpoint selection by validation macro F1 and early stopping rather than use
of the final training epoch.

\section{Results}

\subsection{Backbone Comparison}

Table~\ref{tab:backbone} reports the held-out performance of the six backbones,
each evaluated with ECA under the shared five-fold protocol. DINOv2 ViT-B/14
recorded the highest mean accuracy (91.66\%), macro F1 (90.69\%), and AUC
(96.06\%). Its AUC exceeded EVA02-Base by 0.87 percentage points and
ConvNeXt-Tiny by 0.65 points. ConvNeXtV2-Base produced the highest sensitivity
(88.42\%), whereas EVA02-Base produced the highest specificity (94.80\%).

\begin{table*}[!t]
\caption{Five-fold backbone performance with ECA (mean $\pm$ SD, \%). Best
means are bold.}
\label{tab:backbone}
\centering
\small
\setlength{\tabcolsep}{7pt}
\renewcommand{\arraystretch}{1.12}
\begin{tabular}{lccccc}
\toprule
\textbf{Backbone} & \textbf{Accuracy} & \textbf{Macro F1} &
\textbf{AUC} & \textbf{Sensitivity} & \textbf{Specificity} \\
\midrule
DINOv2 ViT-B/14 & \textbf{91.66$\pm$0.97} & \textbf{90.69$\pm$1.10} &
\textbf{96.06$\pm$0.71} & 87.28$\pm$2.44 & 93.92$\pm$1.05 \\
EVA02-Base & 91.07$\pm$1.64 & 89.90$\pm$1.93 & 95.19$\pm$0.96 &
83.86$\pm$3.75 & \textbf{94.80$\pm$0.95} \\
ConvNeXt-Tiny & 90.23$\pm$1.25 & 89.07$\pm$1.34 & 95.41$\pm$0.60 &
84.44$\pm$2.70 & 93.23$\pm$2.20 \\
ConvNeXtV2-Base & 88.94$\pm$2.08 & 87.96$\pm$2.13 & 95.15$\pm$0.60 &
\textbf{88.42$\pm$2.36} & 89.21$\pm$3.21 \\
BiomedCLIP & 85.58$\pm$1.28 & 84.29$\pm$1.32 & 91.91$\pm$1.26 &
83.49$\pm$1.75 & 86.66$\pm$1.93 \\
MobileNetV3-Large & 61.90$\pm$5.89 & 57.35$\pm$5.14 & 62.03$\pm$6.68 &
44.78$\pm$10.43 & 70.74$\pm$10.81 \\
\bottomrule
\end{tabular}
\end{table*}

\begin{figure*}[t]
  \centering
  \includegraphics[width=0.90\textwidth]{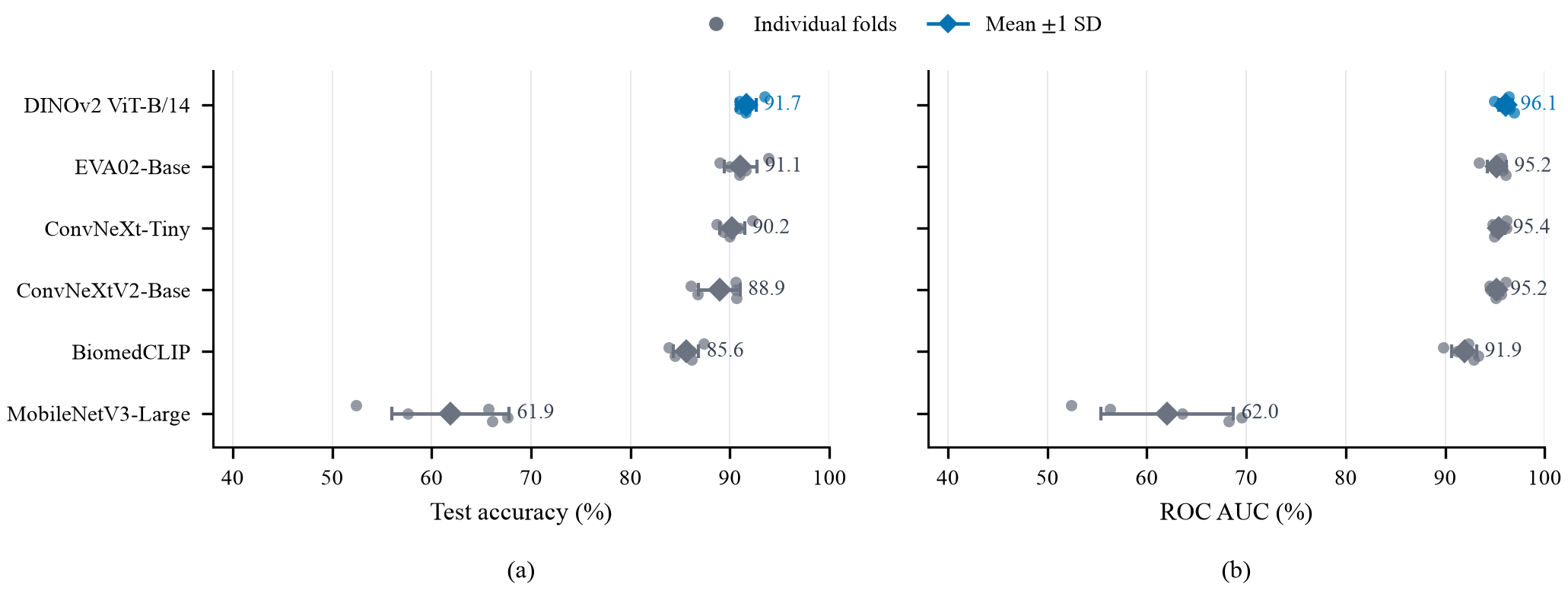}
  \caption{Backbone comparison: (a) test accuracy and (b) ROC AUC. Circles are
  held-out folds; diamonds and bars show mean $\pm$ SD.}
  \label{fig:backbone_comparison}
\end{figure*}

Figure~\ref{fig:backbone_comparison} shows the fold-level accuracy and AUC
distributions. DINOv2 had the highest mean on both plotted metrics, whereas
MobileNetV3-Large was lowest and showed the largest fold-to-fold variation.

\subsection{Representation and Objective Ablation}

Table~\ref{tab:objective_ablation} compares DINOv2 without feature adaptation with DINOv2--ECA under
the evaluated objectives; two center-loss ECA configurations shared the highest
rounded accuracy (91.66\%).

\begin{table*}[!t]
\caption{DINOv2 objective ablation (five-fold mean $\pm$ SD, \%). Best rounded
means are bold.}
\label{tab:objective_ablation}
\centering
\footnotesize
\setlength{\tabcolsep}{4.2pt}
\renewcommand{\arraystretch}{1.12}
\begin{tabular}{lllccccc}
\toprule
\textbf{Adapter} & \textbf{Class loss} & \textbf{Center} & \textbf{Accuracy} &
\textbf{Macro F1} & \textbf{AUC} & \textbf{Sensitivity} & \textbf{Specificity} \\
\midrule
Identity & Focal & Yes & 91.07$\pm$0.67 & 90.09$\pm$0.74 &
95.93$\pm$0.97 & \textbf{87.48$\pm$2.12} & 92.93$\pm$1.18 \\
Identity & Cross-entropy & No & 91.33$\pm$1.66 & 90.32$\pm$1.78 &
95.99$\pm$0.47 & 86.15$\pm$2.03 & 94.02$\pm$2.54 \\
ECA & Focal & Yes & \textbf{91.66$\pm$0.97} & \textbf{90.69$\pm$1.10} &
\textbf{96.06$\pm$0.71} & 87.28$\pm$2.44 & 93.92$\pm$1.05 \\
ECA & Cross-entropy & Yes & \textbf{91.66$\pm$0.62} & 90.57$\pm$0.79 &
95.82$\pm$0.64 & 84.81$\pm$3.43 & \textbf{95.19$\pm$1.34} \\
ECA & Focal & No & 91.40$\pm$1.36 & 90.43$\pm$1.47 &
95.75$\pm$0.73 & 87.29$\pm$1.90 & 93.53$\pm$1.89 \\
ECA & Cross-entropy & No & 91.14$\pm$1.25 & 90.06$\pm$1.42 &
95.70$\pm$0.86 & 85.39$\pm$2.55 & 94.11$\pm$1.27 \\
\bottomrule
\end{tabular}
\end{table*}

ECA improved focal-plus-center accuracy by 0.59 points but reduced
cross-entropy-only accuracy by 0.19 points. All six means lay within 0.59 points,
and the matched McNemar comparison in Fig.~4(a) was not significant
($p=0.368$).

\subsection{Feature-Adapter Comparison}

Adapter accuracies were similar, but ECA recorded the highest mean accuracy and
AUC with five parameters---14,909 times fewer than SE and 154 times fewer
than L0 (Fig.~4(b)--(c)).

\begin{figure*}[!t]
\centering
\begin{minipage}[t]{0.68\textwidth}
\centering
\includegraphics[width=\linewidth]{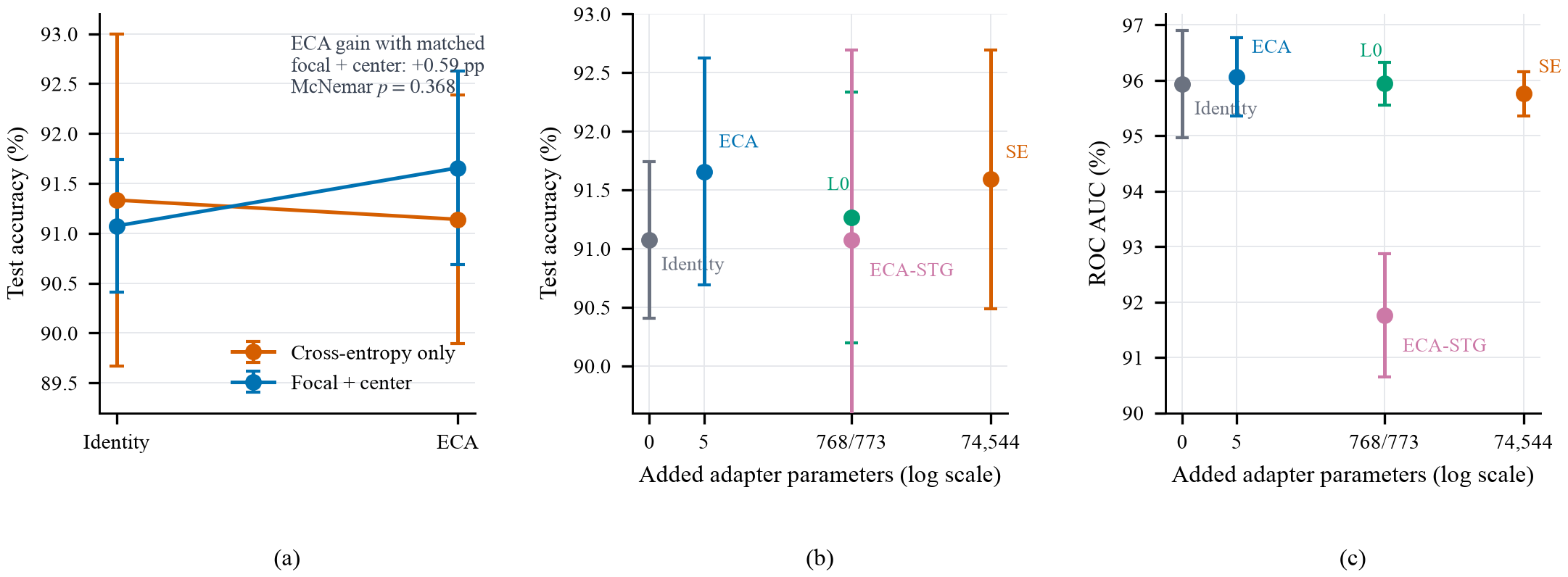}
{\footnotesize Fig.~4.\ Objective and adapter analysis: (a) objective
interaction; (b) accuracy and (c) ROC AUC versus adapter size. Bars show
$\pm1$ SD.\par}
\end{minipage}\hfill
\begin{minipage}[t]{0.29\textwidth}
\centering
\includegraphics[width=0.82\linewidth]{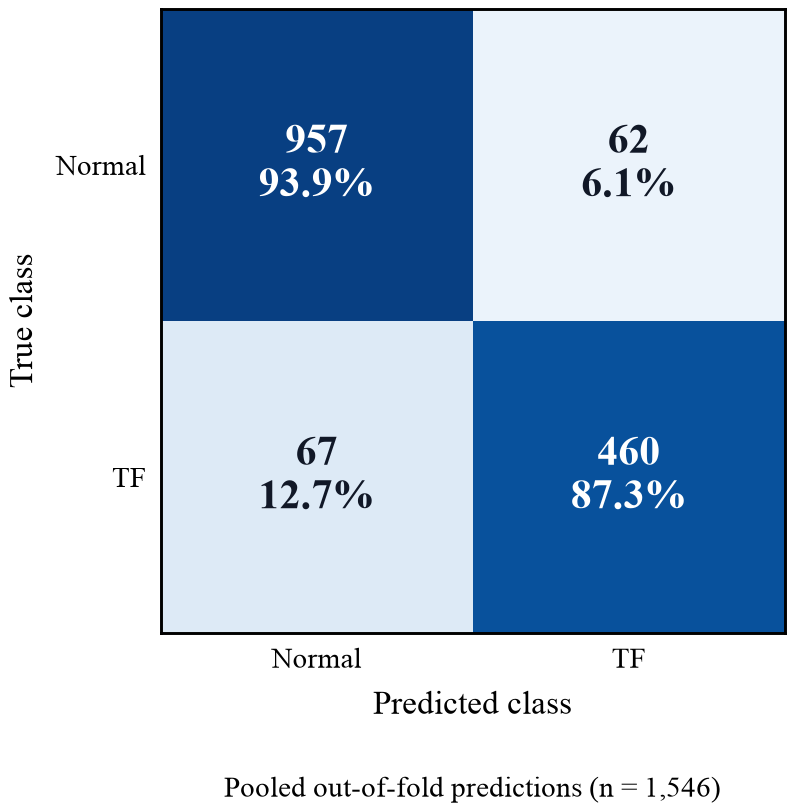}
{\footnotesize Fig.~5.\ Pooled confusion matrix for DINOv2--ECA
($n=1{,}546$); percentages are row-normalized.\par}
\end{minipage}
\end{figure*}

Figure~5 shows 957/1,019 correct Normal and 460/527 correct TF decisions.

\section{Discussion}

Under the common protocol, backbone differences were much larger than adapter
differences, indicating that the fine-tuned DINOv2 representation accounted for most
predictive value. This does not isolate self-supervision as the cause because
architectures, pretraining, resolutions, and optimization responses differed.
ECA's modest observed advantage relative to fold variation supports parameter efficiency,
not decisive superiority; representation sufficiency applies only to the
fine-tuned encoder.

The objective interaction was material: ECA improved focal-plus-center accuracy
by 0.59 points but reduced cross-entropy-only accuracy by 0.19 points. Center loss
with cross-entropy raised specificity to 95.19\% while reducing sensitivity to
84.81\%. Adapter, loss, threshold, and intended clinical cost therefore need
joint selection.

The selected focal-plus-center ECA model achieved 87.28\% sensitivity and 93.92\%
specificity, corresponding to a 12.7\% false-negative rate among TF-positive images. Published
comparisons remain contextual; split definitions and cohorts differ, and larger
field studies provide different evidence. Limitations include a single public
dataset of limited size, excluded TI-only images, unavailable patient identifiers, no external
validation, unaudited SAM~3 masks, one seeded run per fold, and no confidence
intervals for between-model effects. Prospective patient-level validation,
calibration, repeated seeds, and ROI audit are required before deployment.

\FloatBarrier

\section{Conclusion}

DINOv2 ViT-B/14 with ECA achieved 91.66$\pm$0.97\% accuracy,
90.69$\pm$1.10\% macro F1, and 96.06$\pm$0.71\% AUC on the
Normal-versus-TF/TF--TI benchmark. Most observed performance was attributable to the DINOv2
representation; ECA supplied a five-parameter refinement whose benefit depended
on the objective. Future work should evaluate full WHO grading, external
patient-level cohorts, calibration and thresholding, and clinically interpretable
error analysis.

\section*{Code and Data Availability}

Source code for all experiments is publicly available at
\url{https://github.com/HadushHailu/active_trachoma}. The dataset analyzed in
this study is publicly available at Figshare:
\url{https://figshare.com/articles/dataset/TrachomaImages/7551053/1}.

\balance

\end{document}